\documentclass[10pt,twocolumn,letterpaper]{article}

\usepackage{iccv}
\usepackage{times}
\usepackage{epsfig}
\usepackage{graphicx}
\usepackage{amsmath}
\usepackage{amssymb}
\usepackage{amsfonts}
\usepackage{amsfonts}
\usepackage{booktabs}
\usepackage{caption}
\usepackage{subcaption}
\usepackage{algorithmic}
\usepackage{algorithm}

\usepackage{xcolor}

\usepackage{amsthm}
\theoremstyle{definition}

\theoremstyle{proposition}
\newtheorem{proposition}{Proposition}[section]

\usepackage[pagebackref=true,breaklinks=true,letterpaper=true,colorlinks,bookmarks=false]{hyperref}

\iccvfinalcopy 

\def\iccvPaperID{48} 

\ificcvfinal\fi
\begin{document}

\title{No Fuss Distance Metric Learning using Proxies}

\author{Yair Movshovitz-Attias, Alexander Toshev, Thomas K. Leung, Sergey Ioffe, Saurabh Singh\\
Google Research\\
{\tt\small \{yairmov, toshev, leungt, sioffe, saurabhsingh\}@google.com}
}

\maketitle

\begin{abstract}
    We address the problem of distance metric learning (DML), defined as learning a distance consistent with a notion of semantic similarity. Traditionally, for this problem supervision is expressed in the form of sets of points that follow an ordinal relationship -- an anchor point $x$ is similar to a set of positive points $Y$, and dissimilar to a set of negative points $Z$, and a loss defined over these distances is minimized.
    While the specifics of the optimization differ, in this work we collectively call this type of supervision \emph{Triplets} and all methods that follow this pattern \emph{Triplet-Based} methods. These methods are challenging to optimize. A main issue is the need for finding informative triplets, which is usually achieved by a variety of tricks such as increasing the batch size, hard or semi-hard triplet mining, etc. Even with these tricks, the convergence rate of such methods is slow. In this paper we propose to optimize the triplet loss on a different space of triplets, consisting of an anchor data point and similar and dissimilar \textit{proxy} points which are learned as well. These proxies approximate the original data points, so that a triplet loss over the proxies is a tight upper bound of the original loss. This proxy-based loss is empirically better behaved. As a result, the proxy-loss improves on state-of-art results for three standard zero-shot learning datasets, by up to 15\% points, while converging three times as fast as other triplet-based losses.

\end{abstract}

\section{Introduction}
\begin{figure}[t]
  \centering
    \includegraphics[width=0.47\textwidth]{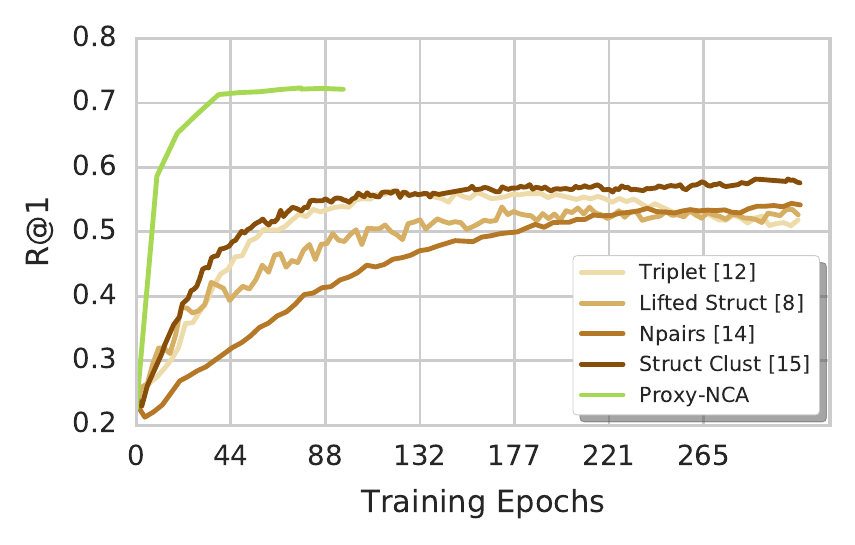}
    \caption{Recall@1 as a function of training step on the Cars196 dataset. Proxy-NCA converges about three times as fast compared with the baseline methods, and results in higher Recall@1 values.}
    \label{fig:recall_step}
\end{figure}

Distance metric learning (DML) is a major tool for a variety of problems in computer vision. It has successfully been employed for image retrieval~\cite{NIPS2016_N-pair}, near duplicate detection~\cite{Zheng_2016_CVPR}, clustering~\cite{hershey2016deep} and zero-shot learning~\cite{oh2016deep}.

A wide variety of formulations have been proposed. Traditionally, these formulations encode a notion of similar and dissimilar data points. For example, \emph{contrastive loss}~\cite{chopra2005learning, hadsell2006dimensionality}, which is defined for a pair of either similar or dissimilar data points. Another commonly used family of losses is \emph{triplet loss}, which is defined by a triplet of data points: an anchor point, and a similar and dissimilar data points. The goal in a triplet loss is to learn a distance in which the anchor point is closer to the similar point than to the dissimilar one.

The above losses, which depend on pairs or triplets of data points, empirically suffer from sampling issues -- selecting informative pairs or triplets is crucial for successfully optimizing them and improving convergence rates. In this work we address this aspect and propose to re-define triplet based losses over a different space of points, which we call proxies. This space approximates the training set of data points (for each data point in the original space there is a proxy point close to it), additionally, it is small enough so that we do not need to sample triplets but can explicitly write the loss over all (or most) of the triplets involving proxies. As a result, this re-defined loss is easier to optimize, and it trains faster (See Figure~\ref{fig:recall_step}). Note that the proxies are learned as part of the model parameters.

In addition, we show that the proxy-based loss is an upper bound to triplet loss and that, empirically, the bound tightness improves as training converges, which justifies the use of proxy-based loss to optimize the original loss.

Further, we demonstrate that the resulting distance metric learning problem has several desirable properties. First and foremost, the obtained metric performs well in the zero-shot scenario, improving state of the art, as demonstrated on three widely used datasets for this problem (CUB200, Cars196 and Stanford Products). Second, the learning problem formulated over proxies exhibits empirically faster convergence than other metric learning approaches. 

\section{Related Work}
There is a large body of work on metric learning, here we focus on its use in computer vision using deep methods. 

An early use of deep methods for metric learning was the introduction of Siamese networks with a
contrastive loss~\cite{chopra2005learning, hadsell2006dimensionality}. Pairs of data points were fed into a network, and the difference between the embeddings produced was used to pull together points from the same class, and push away from each other points from different classes. A shortcoming of this approach is that it can not take directly into account relative distances between classes. Since then, most methods use a notion of \textit{triplets} to provide supervision.

In~\cite{weinberger2006distance} a large margin, nearest neighbor approach is designed to enable k-NN classification. It strives to ensure for each image $x$ a \textit{predefined} set of images from the same class as neighbors that are closer to $x$ than images from other classes with a high separation margin. The set of target neighbors is defined using $l_2$ metric on the input space. The loss function is defined over triplets of points which are sampled during training. 
This sampling becomes prohibiting when the number of classes and training instances becomes large, see Sec~\ref{sub:stochastic-loss} for more details.

To address some of the issues in this and similar work~\cite{schultz2003learning} a Semi-Hard negative mining approach was introduced in~\cite{Schroff_2015_CVPR}. In this
approach, hard triplets were formed by sampling positive/negative instances within a mini-batch with the goal of finding negative examples that are within the margin, but are not too confusing, as those might come from labeling errors in the data. This improved training stability but required large mini-batches - 1800 images in the case of~\cite{Schroff_2015_CVPR}, and training was still slow. Large batches also require non trivial engineering work, e.g. synchronized training with multiple GPUs.

This idea, of incorporating information beyond a single triplet has influenced many approaches. Song et. al.~\cite{oh2016deep} proposed Lifted Structured Embedding, where each positive pair compares the distances against all negative
pairs in the batch weighted by the margin violation.
This provided a smooth loss which incorporates the negative mining functionality.
In~\cite{NIPS2016_N-pair}, the N-Pair Loss was proposed, which used
Softmax cross-entropy loss on pairwise similarity
values within the batch. Inner product is used as a similarity measure between images. The similarity between examples from the same class is encouraged to be higher than the similarity with other images in the batch.
A cluster ranking loss was proposed in~\cite{SongJR016}.
The network first computed the embedding vectors for all images in the batch and ranked a clustering score for the ground truth clustering assignment higher than the clustering score for any other batch assignment with a margin.

Magnet Loss~\cite{rippel2015metric} was designed to compare distributions of classes instead of instances. Each class was represented by a set of $K$ cluster centers, constructed by k-means. In each training iteration, a cluster was sampled, and the M nearest impostor clusters (clusters from different classes) retrieved. From each imposter cluster a set of images were then selected and NCA~\cite{roweis2004neighbourhood} loss used to compare the examples.
Note that, in order to update the cluster assignments, training was paused periodical, and K-Means reapplied.

Our proxy-based approach compares full sets of examples, but both the embeddings and the proxies are trained end-to-end (indeed the proxies are part of the network architecture), without requiring interruption of training to re-compute the cluster centers, or class indices.


\section{Metric Learning using Proxies}\label{sec:metric-learning-using-proxies}
\subsection{Problem Formulation}
We address the problem of learning a distance $d(x,y; \theta)$ between two data points $x$ and $y$. For example, it can be defined as Euclidean distance between embeddings of data obtained via a deep neural net $e(x;\theta)$: $d(x,y; \theta) = ||e(x;\theta) - e(y;\theta)||_2^2$, where $\theta$ are the parameters of the network. To simplify the notation, in the following we drop the full $\theta$ notation, and use $x$ and $e(x;\theta)$ interchangeably.

Often times such distances are learned using similarity style supervision, e.~g.~triplets of similar and dissimilar points (or groups of points) $D=\{(x, y, z)\}$, where in each triplet there is an anchor point $x$, and the second point $y$ (the positive) is more similar to $x$ than the third point $z$ (the negative). Note that both $y$ and, more commonly, $z$ can be sets of positive/negative points. 
We use the notation $Y$, and $Z$ whenever sets of points are used.

The DML task is to learn a distance respecting the similarity relationships encoded in $D$:
\begin{equation}\label{eq:distance}
    d(x,y;\theta) \leq d(x,z;\theta)\quad\textrm{for all}\quad (x,y,z)\in D
\end{equation}

An ideal loss, precisely encoding Eq.~(\ref{eq:distance}), reads:
\begin{equation}
    L_{\textrm{Ranking}}(x,y,z) = H(d(x, y) - d(x, z))
\end{equation}
where $H$ is the Heaviside step function. Unfortunately, this loss is not amenable directly to optimization using stochastic gradient descent as its gradient is zero everywhere. As a result, one resorts to surrogate losses such as Neighborhood Component Analysis (NCA)~\cite{roweis2004neighbourhood} or margin-based triplet loss~\cite{weinberger2006distance, Schroff_2015_CVPR}. 
For example, Triplet Loss uses a hinge function to create a fixed margin between the anchor-positive difference, and the anchor-negative difference:
\begin{equation}\label{eq:triplet_loss}
L_{\textrm{triplet}}(x, y, z) = [d(x, y) + M - d(x, z)]_+
\end{equation}
Where $M$ is the margin, and $[\cdot]_+$ is the hinge function.

Similarly, the NCA loss~\cite{roweis2004neighbourhood} tries to make $x$ closer to $y$ than to any element in a set $Z$ using exponential weighting:
\begin{equation}\label{eq:nca_loss}
    L_{\textrm{NCA}}(x, y, Z) = -\log\left(\frac{\exp(-d(x, y))}{\sum_{z\in Z}\exp(-d(x,z))}\right)
\end{equation}

\subsection{Sampling and Convergence}\label{sub:stochastic-loss}
Neural networks are trained using a form of stochastic gradient descent, where at each optimization step a stochastic loss is formulated by sampling a subset of the training set $D$, called a batch. The size of a batch $b$ is small, e.g. in many modern computer vision network architectures $b=32$. While for classification or regression the loss depends on a single data point from $D$, the above distance learning losses depend on at least three data points, i.e. total number of possible samples could be in $O(n^3)$ for $|D|=n$.

To see this, consider that a common source of triplet supervision is from a classification-style labeled dataset: a triplet $(x,y,z)$ is selected such that $x$ and $y$ have the same label while $x$ and $z$ do not. For illustration, consider a case where points are distributed evenly between $k$ classes. The number of all possible triplets is then $kn/k \cdot ((n/k)-1)(k-1)\cdot n/k =n^2(n-k)(k-1)/k^2=O(n^3)$.

As a result, in metric learning each batch samples a very small subset of all possible triplets, i.e., in the order of $O(b^3)$. Thus, in order to see all triplets in the training one would have to go over $O((n/b)^3)$ steps, while in the case of classification or regression the needed number of steps is $O(n/b)$. Note that $n$ is in the order of hundreds of thousands, while $b$ is between a few tens to about a hundred, which leads to $n/b$ being in the tens of thousands.  

Empirically, the convergence rate of the optimization procedure is highly dependent on being able to see useful triplets, e.g., triplets which give a large loss value as motivated by \cite{Schroff_2015_CVPR}. The authors propose to sample triplets within the data points present in the current batch, this however, does not address the problem of sampling from the whole set of triplets $D$. This is particularly challenging as the number of triplets is so overwhelmingly large. 

\begin{figure}[t]
  \centering
    \includegraphics[width=0.47\textwidth]{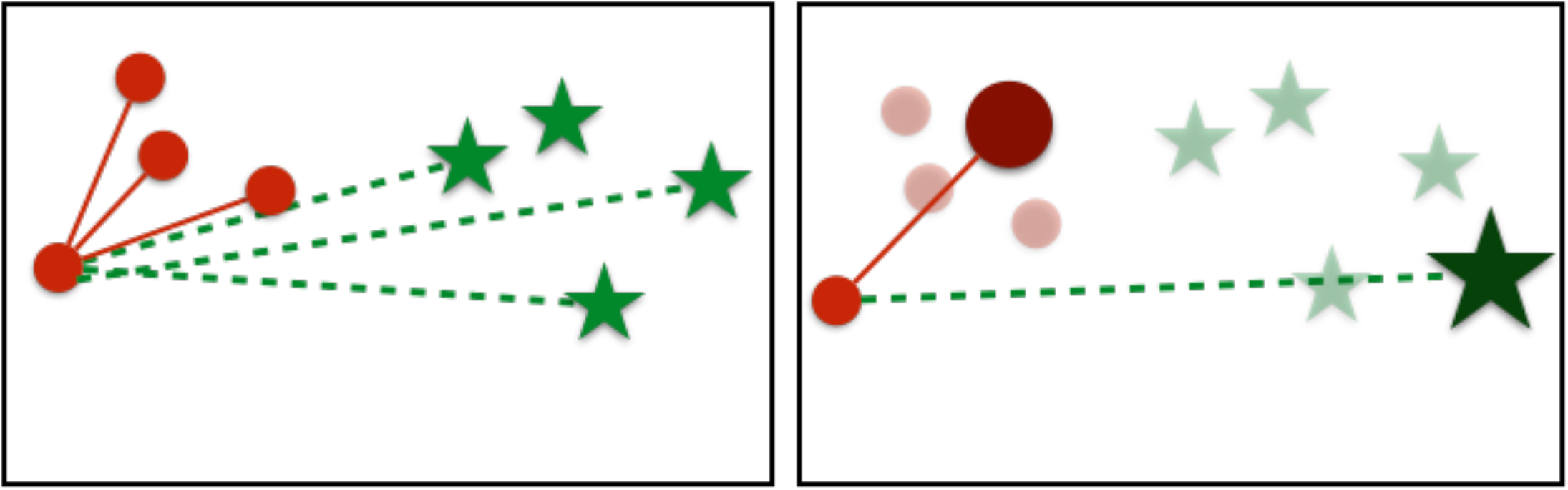}
    \caption{Illustrative example of the power of proxies. [Left panel] There are 48 triplets that can be formed from the instances (small circles/stars). [Right panel] Proxies (large circle/star) serve as a concise representation for each semantic concept, one that fits in memory. By forming triplets using proxies, only 8 comparisons are needed.}
    \label{fig:proxy-illustrative}
\end{figure}

\subsection{Proxy Ranking Loss}\label{sec:ranking_loss}
To address the above sampling problem, we propose to learn a small set of data points $P$ with $|P| \ll |D|$. Intuitively we would like to have $P$ approximate the set of all data points, i.e. for each $x$ there is one element in $P$ which is close to $x$ w.r.t. the distance metric $d$. We call such an element a \emph{proxy} for $x$:
\begin{equation}\label{eq:proxy}
    p(x) = \arg\min_{p \in P}d(x, p)
\end{equation}
and denote the \textit{proxy approximation error} by the worst approximation among all data points
\begin{equation}\label{eq:proxy_precision}
    \epsilon = \max_x d(x, p(x))
\end{equation}

We propose to use these proxies to express the ranking loss, and because the proxy set is smaller than the original training data, the number of triplets would be significantly reduced (see Figure~\ref{fig:proxy-illustrative}). Additionally, since the proxies represent our original data, the reformulation of the loss would implicitly encourage the desired distance relationship in the original training data.

To see this, consider a triplet $(x, y, z)$ for which we are to enforce Eq.~(\ref{eq:distance}). By triangle inequality,
\begin{equation}
|\{d(x,y) - d(x,z)\} - \{d(x,p(y)) - d(x,p(z))\}| \leq 2 \epsilon \nonumber
\end{equation}
As long as $|d(x, p(y)) - d(x, p(z))| > 2\epsilon$, the ordinal relationship between the distance $d(x,y)$ and $d(x,z)$ is not changed when $y, z$ are replaced by the proxies $p(y), p(z)$.  Thus, we can bound the expectation of the ranking loss over the training data:
\begin{align*}
     \mathrm{E}&[L_{\textrm{Ranking}}(x; y, z)] \leq \\
     & \mathrm{E}[L_{\textrm{Ranking}}(x; p(y), p(z))] \, + \\
     & \mathrm{Pr}[|d(x,p(y) - d(x,p(z)| \leq 2 \epsilon]
\end{align*}
    
Under the assumption that all the proxies have norm $\|p\| = N_p$ and all embeddings have the same norm $\|x\| = N_x$, the bound can be tightened. Note that in this case we have, for any $\alpha >0$:
\begin{align*}
    & L_{\textrm{Ranking}}(x,y,z) \\
    & = H(\|\alpha x-p(y)\| - \|\alpha x - p(z)\|) \\
     & = H(\|\alpha x-p(y)\|^2 - \|\alpha x - p(z)\|^2) \\
     & = H(2 \alpha(x^Tp(z) - x^Tp(y))) = H(x^Tp(z) - x^Tp(y)).
\end{align*}
I.e. the ranking loss is scale invariant in $x$.
However, such re-scaling affects the distances between the embeddings and proxies.  We can judiciously choose $\alpha$ to obtain a better bound.  A good value would be one that makes the embeddings and proxies lie on the same sphere, i.e. $\alpha = N_p / N_x$. These assumptions prove easy to satisfy, see Section~\ref{sec:training}.

The ranking loss is difficult to optimize, particularly with gradient based methods.  We argue that many losses, such as NCA loss~\cite{roweis2004neighbourhood}, Hinge triplet loss~\cite{weinberger2006distance}, N-pairs loss~\cite{NIPS2016_N-pair}, etc are merely surrogates for the ranking loss.  In this next section, we show how the proxy approximation can be used to bound the popular NCA loss for distance metric learning.
\section{Training}~\label{sec:training}
In this section we explain how to use the introduced proxies to train a distance based on the NCA formulation. We would like to minimize the total loss, defined as a sum over triplets $(x, y, Z)$ (see Eq.~(\ref{eq:distance})). Instead, we minimize the upper bound, defined as a sum over triplets over an anchor and two proxies $(x, p(y), p(Z))$ (see Eq.~(\ref{eq:total-loss-bound})).

\begin{algorithm}[H]
\begin{algorithmic}
\STATE Randomly init all values in $\theta$ including proxy vectors.
\FOR{$i=1\ldots T$}
    \STATE Sample triplet $(x, y, Z)$ from $D$
    \STATE Formulate proxy triplet $(x, p(y), p(Z))$
    \STATE $l = -\log\left(\frac{\exp(-d(x, p(y)))}{\sum_{p(z)\in p(Z)}\exp(-d(x,p(z)))}\right)$
    \STATE $\theta \gets \theta - \lambda \partial_\theta l$
\ENDFOR
\end{algorithmic}
\caption{Proxy-NCA Training.}
\label{algo:proxy-optimization}
\end{algorithm}

We perform this optimization by gradient descent, as outlined in Algorithm~\ref{algo:proxy-optimization}. At each step, we sample a triplet of a data point and two proxies $(x, p(y), p(z))$, which is defined by a triplet $(x, y, z)$ in the original training data. However, each triplet defined over proxies upper bounds all triplets $(x, y', z')$ whose positive $y'$ and negative $z'$ data points have the same proxies as $y$ and $z$ respectively. This provides convergence speed-up. The proxies can all be held in memory, and sampling from them is simple. 
In practice, when an anchor point is encountered in the batch, one can use its positive proxy as $y$, and \textbf{all} negative proxies as $Z$ to formulate triplets that cover all points in the data.
We back propagate through both points and proxies, and do not need to pause training to re-calculate the proxies at any time.

We train our model with the property that all proxies have the same norm $N_P$ and all embeddings have the norm $N_X$. Empirically such a model performs at least as well as without this constraint, and it makes applicable the tighter bounds discussed in Section~\ref{sec:ranking_loss}. While in the future we will incorporate the equal norm property into the model during training, for the experiments here we simply trained a model with the desired loss, and re-scaled all proxies and embeddings to the unit sphere (note that the transformed proxies are only useful for analyzing the effectiveness of the bounds, and are not used during inference).

\subsection{Proxy Assignment and Triplet Selection} 
In the above algorithm we need to assign the proxies for the positive and negative data points. We experiment with two assignment procedures. 

When triplets are defined by the semantic labels of data points (the positive data point has the same semantic label as the anchor; the negative a different label), then we can associate a proxy with each semantic label: $P=\{p_1\ldots p_L\}$. Let $c(x)$ be the label of $x$. We assign to a data point the proxy corresponding to its label: $p(x) = p_{c(x)}$. We call this \textit{static proxy assignment} as it is defined by the semantic label and does not change during the execution of the algorithm.
Critically, in this case, we no longer need to sample triplets at all. Instead one just needs to sample an anchor point $x$, and use the anchor's proxy as the positive, and the rest as negatives $L_{NCA}(x, p(x), p(Z);\theta)$

In the more general case, however, we might not have semantic labels. Thus, we assign to a point $x$ the closest proxy, as defined in Eq.~(\ref{eq:proxy}). We call this \textit{dynamic proxy assignment} and note that is aligned with the original definition of the term $proxy$. See Section~\ref{sec:evaluation} for evaluation with the two proxy assignment methods.

\subsection{Proxy-based Loss Bound}
In addition to the motivation for proxies in Sec.~\ref{sec:ranking_loss}, we also show in the following that the proxy based surrogate losses upper bound versions of the same losses defined over the original training data. In this way, the optimization of a single triplet of a data point and two proxies bounds a large number of triplets of the original loss. 

More precisely, if a surrogate loss $L$ over triplet $(x, y, z)$ can be bounded by proxy triplet
\begin{equation*}
L(x, y, z) \leq \alpha L(x, p(y), p(z)) + \delta
\end{equation*}
for constant $\alpha$ and $\delta$, then the following bound holds for the total loss:
\begin{equation}\label{eq:total-loss-bound}
    L(D) \leq \frac{\alpha}{|D|}\sum_{x; p_y, p_z\in P}n_{x,p_y,p_z}L(x, p(y), p(z)) + \delta
\end{equation}
where $n_{x,p_y,p_z}=|\{(x, y, z) \in D | p(y)=p_y, p(z) = p_z\}| $
denotes the number of triplets in the training data with anchor $x$ and proxies $p_y$ and $p_z$ for the positive and negative data points.

The quality of the above bound depends on $\delta$, which depends on the loss and as we will see also on the proxy approximation error $\epsilon$.  We will show for concrete loss that the bound gets tighter for small proxy approximation error.

The proxy approximation error depends to a degree on the number of proxies $|P|$. In the extreme case, the number of proxies is equal to the number of data points, and the approximation error is zero. Naturally, the smaller the number of proxies the higher the approximation error. However, the number of terms in the bound is in $O(n|P|^2)$. If $|P| \approxeq n$ then the number of samples needed will again be $O(n^3)$. We would like to keep the number of terms as small as possible, as motivated in the previous section, while keeping the approximation error small as well. Thus, we seek a balance between small approximation error and small number of terms in the loss. In our experiments, the number of proxies varies from a few hundreds to a few thousands, while the number of data points is in the tens/hundreds of thousands.


\paragraph{Proxy loss bounds} For the following we assume that the norms of proxies and data points are constant $|p_x| = N_p$ and $|x| = N_x$, we will denote $\alpha=\frac{1}{N_pN_x}$. Then the following bounds of the original losses by their proxy versions are:
\theoremstyle{proposition}
\begin{proposition}\label{prop:nca}
The NCA loss (see Eq.~(\ref{eq:nca_loss})) is proxy bounded:
\begin{equation*}
    \hat{L}_{\textrm{NCA}}(x, y, Z) \leq \alpha L_{\textrm{NCA}}(x, p_y, p_{Z}) + (1-\alpha)\log(|Z|) + 2\sqrt{2\epsilon}
\end{equation*}
where $\hat{L}_{\textrm{NCA}}$ is defined as $L_{\textrm{NCA}}$ with normalized data points and $|Z|$ is the number of negative points used in the triplet.
\end{proposition}
\theoremstyle{proposition}
\begin{proposition}\label{prop:triplet}
The margin triplet loss (see Eq.~(\ref{eq:triplet_loss})) is proxy bounded:
\begin{equation*}
    \hat{L}_{\textrm{triplet}}(x, y, z) \leq \alpha L_{\textrm{triplet}}(x, p_y, p_z) + (1-\alpha)M + 2\sqrt{\epsilon}
\end{equation*}
where $\hat{L}_{\textrm{triplet}}$ is defined as $L_{\textrm{triplet}}$ with normalized data points.
\end{proposition}
See Appendix for proofs.

\section{Implementation Details}
We used the TensorFlow Deep Learning framework~\cite{abadi2016tensorflow} for all methods described below.
For fair comparison\footnote{We thank the authors of~\cite{SongJR016} for providing their code for the baseline methods, in which we based our model, and for helpful discussions.} we follow the implementation details of~\cite{SongJR016}. We use the Inception~\cite{SzegedyLJSRAEVR14:Inception} architecture with batch normalization~\cite{IoffeS15BatchNorm}.
All methods are first pretrained on ILSVRC 2012-CLS data~\cite{russakovsky2015imagenet}, and then finetuned on the tested datasets. The size of the learned embeddings is set to 64.
The inputs are resized to $256 \times 256$ pixels, and then randomly cropped to $227 \times 227$.
The numbers reported in~\cite{NIPS2016_N-pair} are using multiple random crops during test time, but for fair 
comparison with the other methods, and following the procedure in~\cite{SongJR016}, our implementation uses only
a center crop during test time.
We use the RMSprop optimizer with the margin multiplier constant $\gamma$ decayed at a rate of $0.94$.
The only difference we take from the setup described in~\cite{SongJR016} is that
for our proposed method, we use a batch size $m$ of 32 images (all other methods use $m=128$). We do this to illustrate one
of the benefits of the proposed method - it does not \textit{require} large batches. We have experimentally confirmed that the results are stable when we use larger batch sizes for our method.

Most of our experiments are done with a Proxy-NCA loss. However, proxies can be introduced in many popular metric learning algorithms, as outlined in Section~\ref{sec:metric-learning-using-proxies}. To illustrate this point,
we also report results of using a Proxy-Triplet approach on one of the datasets, see Section~\ref{sec:evaluation} below.

\section{Evaluation}\label{sec:evaluation}
Based on the experimental protocol detailed in~\cite{SongJR016, NIPS2016_N-pair} we evaluate retrieval at $k$ and
clustering quality on data from unseen classes on 3 datasets: CUB200-2011~\cite{wah2011caltech}, Cars196~\cite{krause20133d}, and Stanford Online Products~\cite{oh2016deep}. 
Clustering quality is evaluated using the Normalized Mutual Information 
measure (NMI). NMI is defined as the ratio of the mutual information of the clustering and ground truth, and their harmonic mean.
Let $\Omega = \{\omega_1, \omega_2, \ldots, \omega_k\}$ be the cluster assignments that are, for example, the result of K-Means clustering. That is, $\omega_i$ contains the instances assigned to the i'th cluster.
Let $\mathbb{C} = \{c_1, c_2, \ldots, c_m\}$ be the ground truth classes, where $c_j$ contains the instances from class $j$.
\begin{equation}
    \text{NMI}(\Omega, \mathbb{C}) = 2\frac{\text{I}(\Omega, \mathbb{C})}{H(\Omega) + H(\mathbb{C})}.
\end{equation}
Note that NMI is invariant to label permutation which is a desirable property for for our evaluation. For more information on clustering quality measurement see~\cite{manning2008introduction}.

We compare our Proxy-based method with 4 state-of-the-art deep metric learning approaches: Triplet Learning with semi-hard negative mining~\cite{Schroff_2015_CVPR}, Lifted Structured Embedding~\cite{oh2016deep}, the N-Pairs deep metric 
loss~\cite{NIPS2016_N-pair}, and Learnable Structured Clustering~\cite{SongJR016}. In all our experiments we use the same data splits as~\cite{SongJR016}.

\subsection{Cars196}
\begin{figure}[t]
  \centering
    \includegraphics[width=0.47\textwidth]{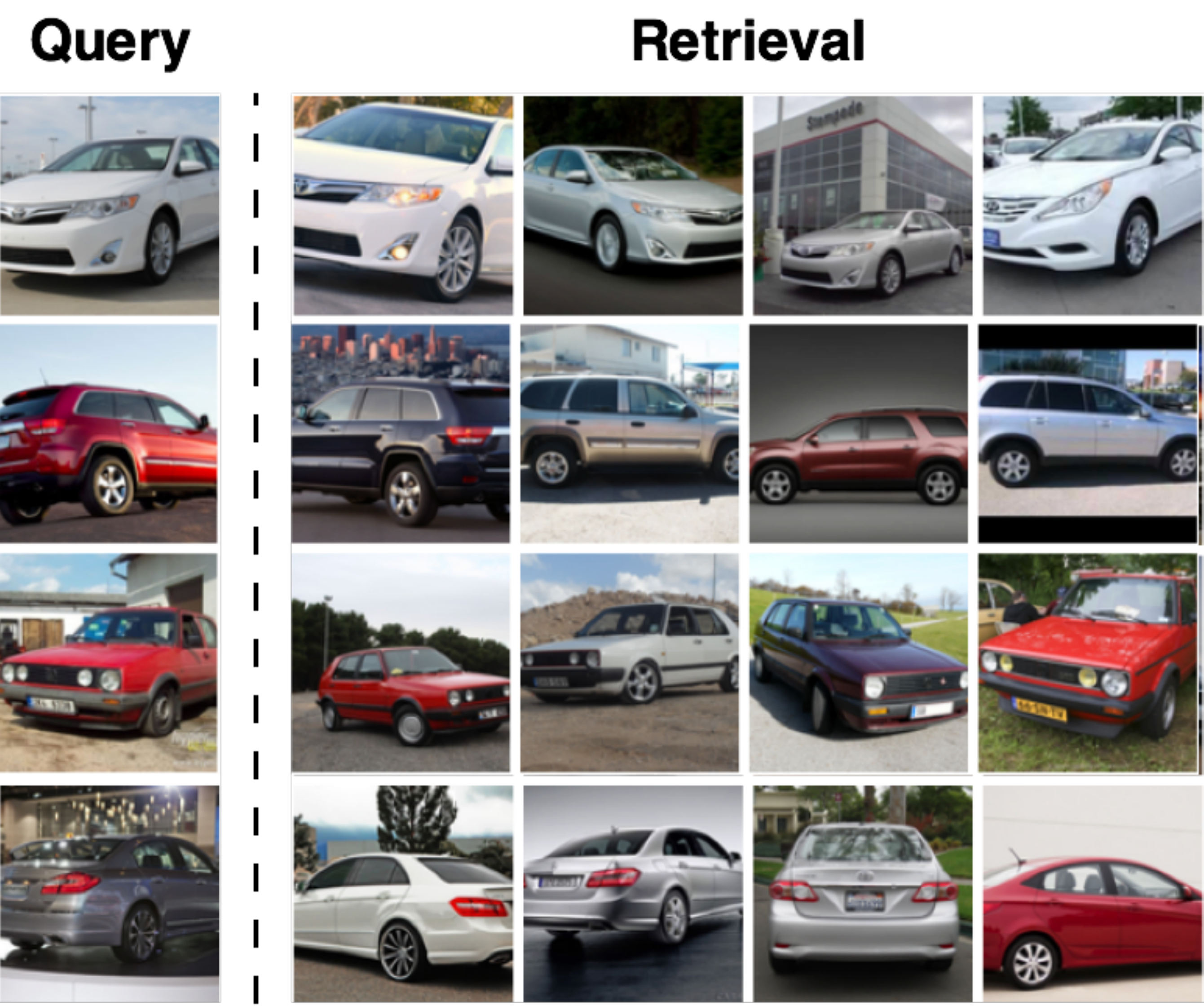}
    \caption{Retrieval results on a set of images from the Cars196 dataset using our proposed proxy-based training method. Left column contains query images. The results are ranked by distance.}
    \label{fig:cars196-retrieval}
\end{figure}

The Cars196 dataset~\cite{krause20133d} is a fine-grained car category dataset containing 16,185 images of 196 car models. Classes are at the level of \emph{make}-\emph{model}-\emph{year}, for example, Mazda-3-2011. In our experiments we split the dataset such that 50\% of the \textit{classes} are used for training, and 50\% are used for evaluation. Table~\ref{tab:cars196} shows recall-at-k and NMI scores for all methods on the Cars196 dataset. Proxy-NCA has a 15 percentage points (26\% relative) improvement in recall@1 from previous state-of-the-art, and a 6\% point gain in NMI.
Figure~\ref{fig:cars196-retrieval} shows example retrieval results on the test set of the Cars196 dataset.

\begin{table}[t]
\small
    \centering
    \begin{tabular}{lccccc}
        \toprule
        {}                                        &   R@1  &   R@2  &   R@4  &   R@8  &  NMI \\
        \midrule
        Triplet Semihard~\cite{Schroff_2015_CVPR} &  51.54 &  63.78 &  73.52 &  81.41 &  53.35 \\
        Lifted Struct~\cite{oh2016deep}           &  52.98 &  66.70 &  76.01 &  84.27 &  56.88 \\
        Npairs~\cite{NIPS2016_N-pair}             &  53.90 &  66.76 &  77.75 &  86.35 &  57.79 \\
        Proxy-Triplet                             &  55.90 &  67.99 &  74.04 &  77.95 &  54.44 \\
        Struct Clust~\cite{SongJR016}             &  58.11 &  70.64 &  80.27 &  87.81 &  59.04 \\
        Proxy-NCA                                 &  \textbf{73.22} &  \textbf{82.42} &  \textbf{86.36} &  \textbf{88.68} &  \textbf{64.90}\\
        \bottomrule
    \end{tabular}
    \caption{Retrieval and Clustering Performance on the Cars196 dataset. Bold indicates best results.}
    \label{tab:cars196}
\end{table}

\subsection{Stanford Online Products dataset}\label{sec:stanford-products}
\begin{figure}[t]
  \centering
    \includegraphics[width=0.47\textwidth]{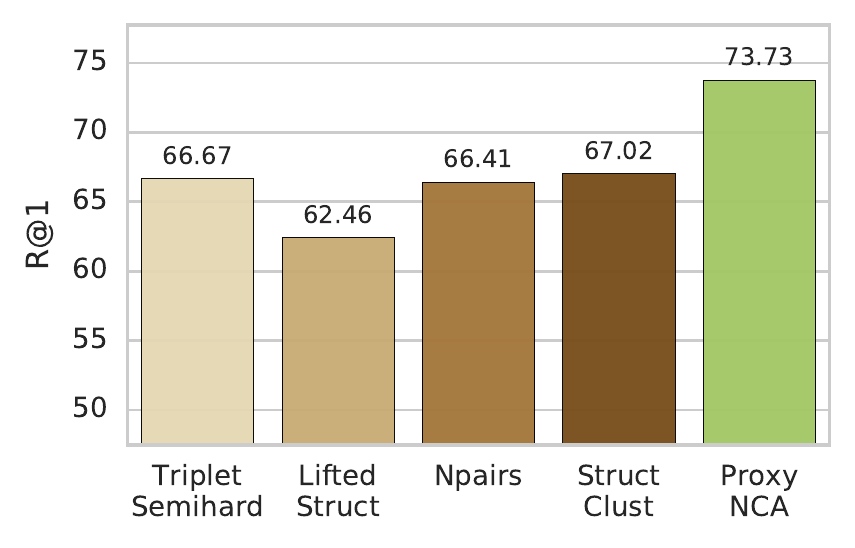}
    \caption{Recall@1 results on the Stanford Product Dataset. Proxy-NCA has a 6\% point gap with previous SOTA.}
    \label{fig:stanford-product-recall}
\end{figure}

The Stanford product dataset contains 120,053 images of 22,634 products downloaded from eBay.com.
For training, 59,5511 out of 11,318 classes are used, and 11,316 classes (60,502 images) are held out for testing.
This dataset is more challenging as each product has only about 5 images, and at first seems well suited for tuple-sampling approaches, and less so for our proxy formulation. Note that holding in memory 
11,318 float proxies of dimension 64 takes less than 3Mb. Figure~\ref{fig:stanford-product-recall} shows recall-at-1 results on this dataset. Proxy-NCA has over a 6\% gap from previous state of the art. 
Proxy-NCA compares favorably on clustering as well, with a score of 90.6. This, compared with the top method, described in~\cite{SongJR016} which has an NMI score of 89.48. The difference is statistically significant. 


\begin{figure}[t]
  \centering
    \includegraphics[width=0.47\textwidth]{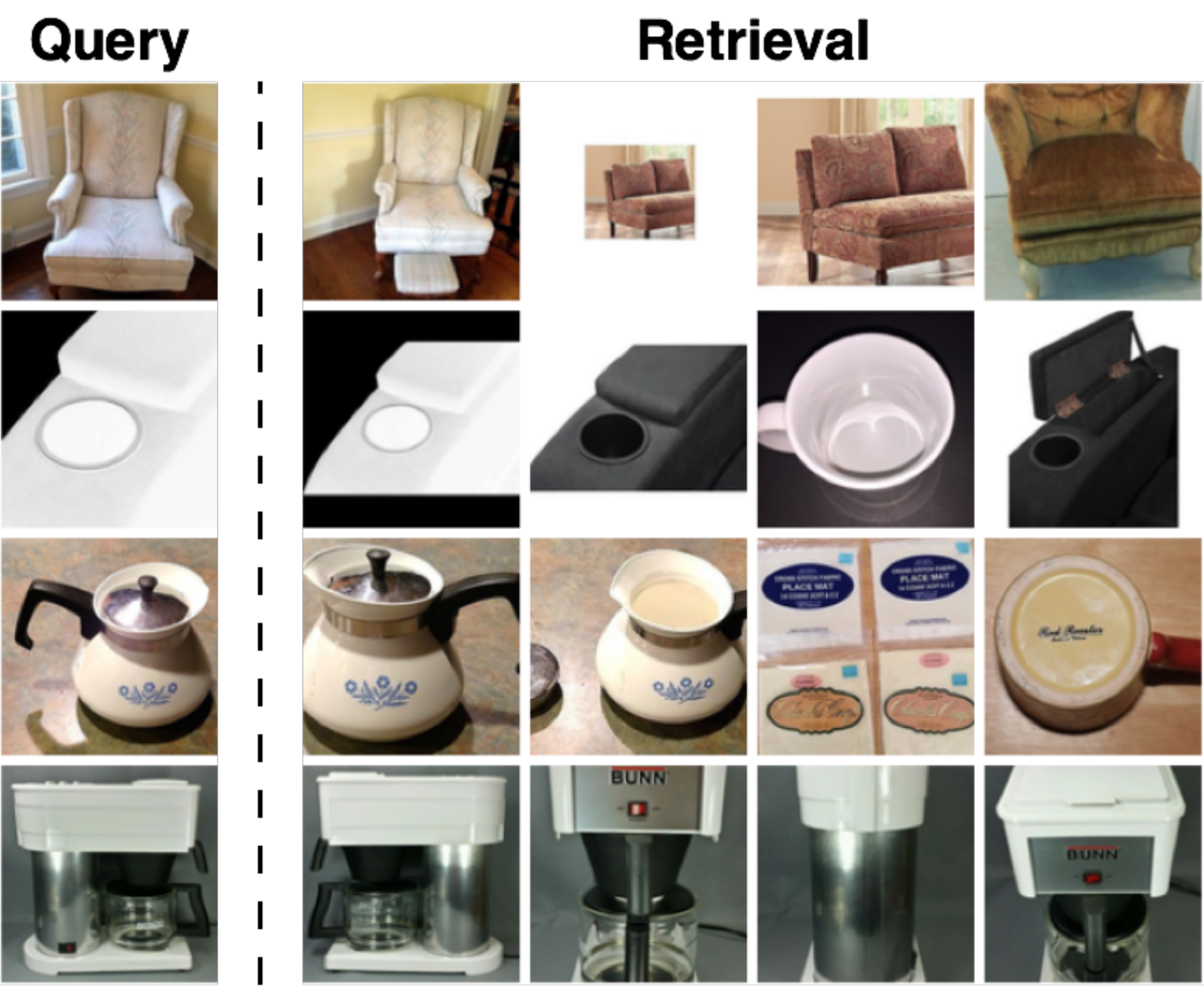}
    \caption{Retrieval results on a randomly selected set of images from the Stanford Product dataset. Left column contains query images. The results are ranked by distance. Note the rotation invariance exhibited by the embedding.}
    \label{fig:stanford-product-retrieval}
\end{figure}

Figure~\ref{fig:stanford-product-retrieval} shows example retrieval results on images from the Stanford Product dataset. Interestingly, the embeddings show a high degree of rotation invariance.

\subsection{CUB200}
The Caltech-UCSD Birds-200-2011 dataset contains 11,788 images of birds from 200 classes of fine-grained bird species. We use the first 100 classes as training data for the metric learning methods, and the remaining 100 classes for evaluation. Table~\ref{tab:cub200} compares the proxy-NCA with the baseline methods. Birds are notoriously hard to classify, as the inner-class variation is quite large when compared to the initra-class variation. This is apparent when observing the results in the table. All methods perform less well than in the other datasets. Proxy-NCA improves on SOTA for recall at 1-2 and on the clustering metric.

\begin{table}[t]
\small
    \centering
    \begin{tabular}{lccccc}
        \toprule
        {}                                        &    R@1 &    R@2 &    R@4 &    R@8 &   NMI \\
        \midrule
        Triplet Semihard~\cite{Schroff_2015_CVPR} &  42.59 &  55.03 &  66.44 &  77.23 &  55.38 \\
        Lifted Struct~\cite{oh2016deep}           &  43.57 &  56.55 &  68.59 &  79.63 &  56.50 \\
        Npairs~\cite{NIPS2016_N-pair}             &  45.37 &  58.41 &  69.51 &  79.49 &  57.24 \\
        Struct Clust~\cite{SongJR016}             &  48.18 &  61.44 &  \textbf{71.83} &  \textbf{81.92} & 59.23 \\
        Proxy NCA                                 &  \textbf{49.21} &  \textbf{61.90} &  67.90 &  72.40 & \textbf{59.53}\\
        \bottomrule
    \end{tabular}
    \caption{Retrieval and Clustering Performance on the CUB200 dataset.}
    \label{tab:cub200}
\end{table}

\subsection{Convergence Rate}
The tuple sampling problem that affects most metric learning methods makes them slow to train. By keeping all proxies in memory we eliminate the need for sampling tuples, and mining for hard negative to form tuples. Furthermore, the proxies act as a memory that persists between batches. This greatly speeds up learning.
Figure~\ref{fig:recall_step} compares the training speed of all methods on the Cars196 dataset. Proxy-NCA trains much faster than other metric learning methods, and converges about three times as fast.

\subsection{Fractional Proxy Assignment}
\begin{figure}[t]
  \centering
    \includegraphics[width=0.47\textwidth]{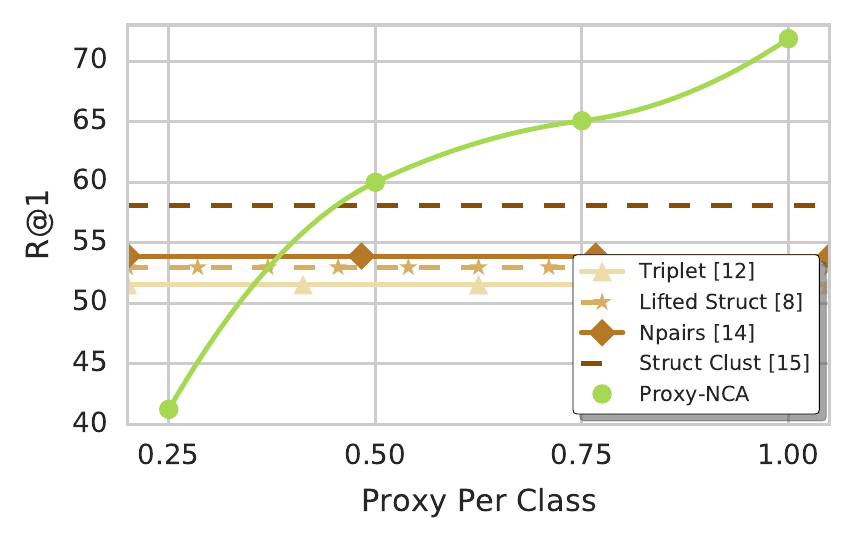}
    \caption{Recall@1 results as a function of ratio of proxies to semantic labels. When allowed 0.5 proxies per label or more, Proxy-NCA compares favorably with previous state of the art.}
    \label{fig:ppc-recall}
\end{figure}

Metric learning requires learning from a large set of semantic labels at times. Section~\ref{sec:stanford-products} shows an example of such a large label set. Even though Proxy-NCA works well in that instance, and the memory footprint of the proxies is small, here we examine the case where one's computational budget does not allow a one-to-one assignment of proxies to semantic labels.
Figure~\ref{fig:ppc-recall} shows the results of an experiment in which we vary the ratio of labels to proxies on the Cars196 dataset. We modify our \textit{static} proxy assignment method to randomly pre-assign semantic labels to proxies. If the number of proxies is smaller than the number of labels, multiple labels are assigned to the same proxy. So in effect each semantic label has influence on a fraction of a proxy.
Note that when $\text{proxy-per-class} \geq 0.5$ Proxy-NCA has better performance than previous methods.

\subsection{Dynamic Proxy Assignment}
\begin{figure}[t]
  \centering
    \includegraphics[width=0.47\textwidth]{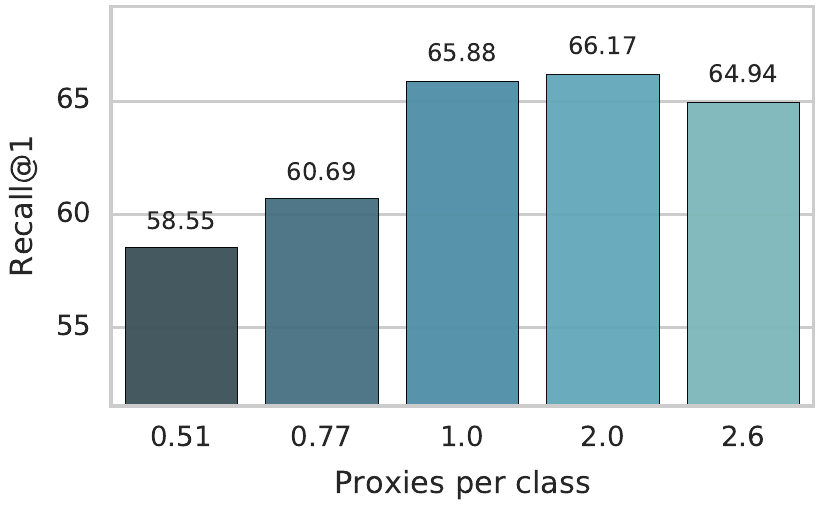}
    \caption{Recall@1 results for dynamic assignment on the Cars196 dataset as a function of proxy-to-semantic-label ratio. More proxies allow for better fitting of the underlying data, but one needs to be careful to avoid over-fitting.}
    \label{fig:dynamic:num-proxies}
\end{figure}

In many cases, the assignment of triplets, i.e. selection of a positive, and negative example to use with the anchor instance, is based on the use of a semantic concept -- two images of a dog need to be more similar than an image of a dog and an image of a cat. These cases are easily handled by our static proxy assignment, which was covered in the experiments above. In some cases however, there are no semantic concepts to be used, and a dynamic proxy assignment is needed. In this section we show results using this assignment scheme.
Figure~\ref{fig:dynamic:num-proxies} shows recall scores for the Cars196 dataset using the dynamic assignment.
The optimization becomes harder to solve, specifically due to the non-differentiable argmin term in Eq.(\ref{eq:proxy}). However, it is interesting to note that first, a budget of $0.5$ proxies per semantic concept is again enough to improve on state of the art, and one does see some benefit of expanding the proxy budget beyond the number of semantic concepts.

\section{Discussion}
In this paper we have demonstrated the effectiveness of using proxies for the task of deep metric learning.
Using proxies, which are saved in memory and trained using back-prop, training time is reduced, and the resulting models achieve a new state of the art. We have presented two proxy assignment schemes -- a static one, which can be used when semantic label information is available, and a dynamic one which is used when the only supervision comes in the form of similar and dissimilar triplets. Furthermore, we show that a loss defined using proxies, upper bounds the original, instance-based loss. If the proxies and instances have constant norms, we show that a well optimized proxy-based model does not change the ordinal relationship between pairs of instances. 

Our formulation of Proxy-NCA loss produces a loss very similar to the standard cross-entropy loss used in classification.  However, we arrive at our formulation from a different direction: we are not interested in the actual classifier and indeed discard the proxies once the model has been trained.  Instead, the proxies are auxiliary variables, enabling more effective optimization of the embedding model parameters.  As such, our formulation not only enables us to surpass the state of the art in zero-shot learning, but also offers an explanation to the effectiveness of the standard trick of training a classifier, and using its penultimate layer's output as the embedding.


\section*{Acknowledgments}
The authors would like to thank Hossein Mobahi, Zhen Li, Hyun Oh Song, Vincent Vanhoucke, and Florian Schroff for helpful discussions.

\section*{Appendix}
\textit{Proof of Proposition~4.1:} In the following for a vector $x$ we will denote its unit norm vector by $\hat{x}=x/|x|$. 

First, we can upper bound the dot product of a unit normalized data points $\hat{x}$ and $\hat{y}$ by the dot product of unit normalized point $\hat{x}$ and proxy $\hat{p_y}$ using the Cauchy inequality as follows:
\begin{equation}
    \hat{x}^T(\hat{z}-\hat{p_z}) \leq |\hat{x}||\hat{z}-\hat{p_z}| \\
    \leq \sqrt{\epsilon}
\end{equation}

Hence:
\begin{equation}\label{eq:bound1}
    \hat{x}^T\hat{z}\leq\hat{x}^T\hat{p_z} + \sqrt{\epsilon}\\
\end{equation}
Similarly, one can obtain an upper bound for the negative dot product:
\begin{equation}\label{eq:bound2}
    -\hat{x}^T\hat{y}\leq -\hat{x}^T\hat{p_y} + \sqrt{\epsilon}\\
\end{equation}
Using the above two bounds we can upper bound the original NCA loss $L_{\textrm{NCA}}(\hat{x}, \hat{y}, Z)$:
\begin{align}
    & =-\log\left(\frac{\exp(-1/2|\hat{x}-\hat{y}|^2)}{\sum_{z\in Z}\exp(-1/2|\hat{x}-\hat{z}|^2)}\right) \nonumber\\
    & = -\log\left(\frac{\exp(-1+\hat{x}^T\hat{y})}{\sum_{z\in Z}\exp(-1+\hat{x}^T\hat{z})}\right) \nonumber\\
    & = -\log\left(\frac{\exp(\hat{x}^T\hat{y})}{\sum_{z\in Z}\exp(\hat{x}^T\hat{z})}\right) \nonumber\\
    & = -\hat{x}^T\hat{y} + \log(\sum_{z\in Z}\exp(\hat{x}^T\hat{z})) \\
    & \leq -\hat{x}^T\hat{p_y} + \sqrt{\epsilon} + \log(\sum_{z\in Z}\exp(\hat{x}^T\hat{p_z} + \sqrt{\epsilon})) \nonumber\\
    & = -\hat{x}^T\hat{p_y} + \log(\sum_{z\in Z}\exp(\hat{x}^T\hat{p_z}))  + 2\sqrt{\epsilon} \nonumber\\
    & = L_{\textrm{NCA}}(\hat{x}, \hat{p_y}, \hat{p_Z})  + 2\sqrt{\epsilon} \label{eq:bound_step_1}
\end{align}

Further, we can upper bound the above loss of unit normalized vectors by a loss of unnormalized vectors. For this we would make the assumption, which empirically we have found true, that for all data points $|x| = N_x > 1$. In practice these norm are much larger than 1.

Lastly, if we denote by $\beta = \frac{1}{N_xN_p}$ and under the assumption that $\beta < 1$, we can apply the following version of the Hoelder inequality defined for positive real numbers $a_i$:
\begin{equation*}
\sum_{i=1}^n a_i^\beta \leq n^{1-\beta}(\sum_{i=1}^n a_i)^\beta     
\end{equation*}
to upper bound the sum of exponential terms:
\begin{align*}
    &\sum_{z\in Z}\exp(\hat{x}^T\hat{p_z})  = \sum_{z\in Z}\exp(\beta x^Tp_z)\\
    & = \sum_{z\in Z}\exp(x^Tp_z)^\beta \leq |Z|^{1-\beta} (\sum_{z\in Z}\exp(x^Tp_z))^\beta
\end{align*}
Hence, the above loss $L_{\textrm{NCA}}$ with unit normalized points is bounded as:
\begin{align}
    & L_{\textrm{NCA}}(\hat{x}, \hat{p_y}, \hat{p_Z}) \nonumber\\
    & \leq -\frac{x^Tp_y}{|x||p_y|} + \log(|Z|^{1-\beta} (\sum_{z\in Z}\exp(x^Tp_z))^\beta) \nonumber\\
    & = -\beta {x^Tp_y} + \beta\log(\sum_{z\in Z}\exp(x^Tp_z)) + \log(|Z|^{1-\beta}) \nonumber\\
    & = {\frac{\beta}{2}|x-p_y|^2} + \beta\log(\sum_{z\in Z}\exp(-\frac{1}{2}|x-p_z|^2)) + \log(|Z|^{1-\beta}) \nonumber\\
    & = \beta L_{\textrm{NCA}}(x, p_y, p_Z) + (1-\beta)\log(|Z|) \label{eq:bound_step_2}
\end{align}
for $\beta=\frac{1}{N_xN_p}$. The propositions follows from Eq.~(\ref{eq:bound_step_1}) and Eq.~(\ref{eq:bound_step_2}).
\qed

\hfill \break\textit{Proof Proposition~4.2:} We will bound the term inside the hinge function in Eq.~(\ref{eq:triplet_loss}) for normalized data points using the bounds (\ref{eq:bound1}) and (\ref{eq:bound2}) from previous proof:
\begin{align*}
    & |\hat{x}-\hat{y}|^2 - |\hat{x}-\hat{z}|^2 + M = -2\hat{x}^T\hat{y} + 2\hat{x}^T\hat{z} + M \\
    & \leq -2\hat{x}^T\hat{p_y} + 2\hat{x}^T\hat{p_z} + 2\sqrt{\epsilon} + M \\
\end{align*}
Under the assumption that the data points and the proxies have constant norms, we can convert the above dot products to products of unnormalized points:
\begin{align*}
    & -2\hat{x}^T\hat{p_y} + 2\hat{x}^T\hat{p_z} + 2\sqrt{\epsilon}  + M \\
    & = \alpha(-2x^Tp_y + 2x^Tp_z)  + 2\sqrt{\epsilon}  + M \\
    & =\alpha(|x-p_y|^2 - |x-p_z|^2)  + 2\sqrt{\epsilon}  + M \\
    & =\alpha(|x-p_y|^2 - |x-p_z|^2 + M)  +  (1-\alpha)M + 2\sqrt{\epsilon} \\
\end{align*}
\qed


{\small
\bibliographystyle{ieee}
\bibliography{refs}

\begin{thebibliography}{10}\itemsep=-1pt

\bibitem{abadi2016tensorflow}
M.~Abadi, A.~Agarwal, P.~Barham, E.~Brevdo, Z.~Chen, C.~Citro, G.~S. Corrado,
  A.~Davis, J.~Dean, M.~Devin, et~al.
\newblock Tensorflow: Large-scale machine learning on heterogeneous distributed
  systems.
\newblock {\em arXiv preprint arXiv:1603.04467}, 2016.

\bibitem{chopra2005learning}
S.~Chopra, R.~Hadsell, and Y.~LeCun.
\newblock Learning a similarity metric discriminatively, with application to
  face verification.
\newblock In {\em Computer Vision and Pattern Recognition, 2005. CVPR 2005.
  IEEE Computer Society Conference on}, 2005.

\bibitem{hadsell2006dimensionality}
R.~Hadsell, S.~Chopra, and Y.~LeCun.
\newblock Dimensionality reduction by learning an invariant mapping.
\newblock In {\em Computer vision and pattern recognition, 2006 IEEE computer
  society conference on}, volume~2, pages 1735--1742. IEEE, 2006.

\bibitem{hershey2016deep}
J.~R. Hershey, Z.~Chen, J.~Le~Roux, and S.~Watanabe.
\newblock Deep clustering: Discriminative embeddings for segmentation and
  separation.
\newblock In {\em Acoustics, Speech and Signal Processing (ICASSP), 2016 IEEE
  International Conference on}, pages 31--35. IEEE, 2016.

\bibitem{IoffeS15BatchNorm}
S.~Ioffe and C.~Szegedy.
\newblock Batch normalization: Accelerating deep network training by reducing
  internal covariate shift.
\newblock {\em arXiv preprint arXiv: 1502.03167}, 2015.

\bibitem{krause20133d}
J.~Krause, M.~Stark, J.~Deng, and L.~Fei-Fei.
\newblock 3d object representations for fine-grained categorization.
\newblock In {\em Proceedings of the IEEE International Conference on Computer
  Vision Workshops}, pages 554--561, 2013.

\bibitem{manning2008introduction}
C.~D. Manning, P.~Raghavan, H.~Sch{\"u}tze, et~al.
\newblock {\em Introduction to information retrieval}.
\newblock Cambridge university press Cambridge, 2008.

\bibitem{oh2016deep}
H.~Oh~Song, Y.~Xiang, S.~Jegelka, and S.~Savarese.
\newblock Deep metric learning via lifted structured feature embedding.
\newblock In {\em Proceedings of the IEEE Conference on Computer Vision and
  Pattern Recognition}, 2016.

\bibitem{rippel2015metric}
O.~Rippel, M.~Paluri, P.~Dollar, and L.~Bourdev.
\newblock Metric learning with adaptive density discrimination.
\newblock {\em arXiv preprint arXiv:1511.05939}, 2015.

\bibitem{roweis2004neighbourhood}
S.~Roweis, G.~Hinton, and R.~Salakhutdinov.
\newblock Neighbourhood component analysis.
\newblock {\em Adv. Neural Inf. Process. Syst.(NIPS)}, 2004.

\bibitem{russakovsky2015imagenet}
O.~Russakovsky, J.~Deng, H.~Su, J.~Krause, S.~Satheesh, S.~Ma, Z.~Huang,
  A.~Karpathy, A.~Khosla, M.~Bernstein, et~al.
\newblock Imagenet large scale visual recognition challenge.
\newblock {\em International Journal of Computer Vision}, 115(3):211--252,
  2015.

\bibitem{Schroff_2015_CVPR}
F.~Schroff, D.~Kalenichenko, and J.~Philbin.
\newblock Facenet: A unified embedding for face recognition and clustering.
\newblock In {\em The IEEE Conference on Computer Vision and Pattern
  Recognition (CVPR)}, June 2015.

\bibitem{schultz2003learning}
M.~Schultz and T.~Joachims.
\newblock Learning a distance metric from relative comparisons.
\newblock In {\em NIPS}, volume~1, page~2, 2003.

\bibitem{NIPS2016_N-pair}
K.~Sohn.
\newblock Improved deep metric learning with multi-class n-pair loss objective.
\newblock In D.~D. Lee, M.~Sugiyama, U.~V. Luxburg, I.~Guyon, and R.~Garnett,
  editors, {\em Advances in Neural Information Processing Systems 29}, pages
  1857--1865. Curran Associates, Inc., 2016.

\bibitem{SongJR016}
H.~O. Song, S.~Jegelka, V.~Rathod, and K.~Murphy.
\newblock Learnable structured clustering framework for deep metric learning.
\newblock {\em The IEEE Conference on Computer Vision and Pattern Recognition
  (CVPR)}, 2017.

\bibitem{SzegedyLJSRAEVR14:Inception}
C.~Szegedy, W.~Liu, Y.~Jia, P.~Sermanet, S.~E. Reed, D.~Anguelov, D.~Erhan,
  V.~Vanhoucke, and A.~Rabinovich.
\newblock Going deeper with convolutions.
\newblock {\em arXiv preprint arXiv: 1409.4842}, 2014.

\bibitem{wah2011caltech}
C.~Wah, S.~Branson, P.~Welinder, P.~Perona, and S.~Belongie.
\newblock The caltech-ucsd birds-200-2011 dataset.
\newblock 2011.

\bibitem{weinberger2006distance}
K.~Q. Weinberger, J.~Blitzer, and L.~Saul.
\newblock Distance metric learning for large margin nearest neighbor
  classification.
\newblock {\em Advances in neural information processing systems}, 18:1473,
  2006.

\bibitem{Zheng_2016_CVPR}
S.~Zheng, Y.~Song, T.~Leung, and I.~Goodfellow.
\newblock Improving the robustness of deep neural networks via stability
  training.
\newblock In {\em The IEEE Conference on Computer Vision and Pattern
  Recognition (CVPR)}, June 2016.

\end{thebibliography}
}

\end{document}